\documentclass[10pt, a4paper, conference, compsocconf]{IEEEtran}
\ifCLASSINFOpdf
\else
\fi

\usepackage{graphicx}
\usepackage{amsmath}
\usepackage{amssymb}
\usepackage{breqn}

\begin{document}
%
\title{Common Representation Learning Using Step-based Correlation Multi-Modal CNN}


\author{\IEEEauthorblockN{Gaurav Bhatt, Piyush Jha, and Balasubramanian Raman}
\IEEEauthorblockA{Indian Institute of Technology Roorkee\\
Roorkee, India.\\
\{gauravbhatt.cs.iitr,piyushnit15\}@gmail.com, balarfma@iitr.ac.in}
}


%


\maketitle

\begin{abstract}
Deep learning techniques have been successfully used in learning a common representation for multi-view data, wherein the different modalities are projected onto a common subspace. In a broader perspective, the techniques used to investigate common representation learning falls under the categories of canonical correlation-based approaches and autoencoder based approaches. In this paper, we investigate the performance of deep autoencoder based methods on multi-view data. We propose a novel step-based correlation multi-modal CNN (CorrMCNN) which reconstructs one view of the data given the other while increasing the interaction between the representations at each hidden layer or every intermediate step. Finally, we evaluate the performance of the proposed model on two benchmark datasets - MNIST and XRMB. Through extensive experiments, we find that the proposed model achieves better performance than the current state-of-the-art techniques on joint common representation learning and transfer learning tasks.
\end{abstract}


%
\IEEEpeerreviewmaketitle
\section{Introduction}

Representation of data in multiple views can be seen in applications related to machine learning, computer vision, and natural language processing. At times it is beneficial to combine different modalities of data since an amalgamation of multiple views is likely to capture more meaningful information than a representation that is fit for only a specific modality. For example, consider the task of abstract scene recognition in a movie \cite{rasheed2003scene} with text annotations as labels. Movie data is comprised of video frames (images) along with audio. Here, images and audio are two different representations of same data with different representative features. Thus, combining these two modalities into a common subspace can help in the task of classification of abstract scenes from videos. Similarly, multi-view data have been used in audio + articulation \cite{arora2012kernel,wang2015unsupervised}, images + text \cite{vinyals2017show,you2016image}, training transliterated corpora (bilingual data) \cite{hermann2014multilingual,klementiev2012inducing,hermann2013multilingual,chandar2016correlational}, or bridge autoencoders \cite{saha2016correlational}.

The techniques used for common representation learning (CRL) of multi-view data can be categorized into two categories - canonical based approaches and autoencoder based methods. Canonical correlation analysis (CCA) \cite{hardle2007canonical,thompson2005canonical} is matrix factorization method that maximizes the correlation between two views of data by projecting different modalities onto a common subspace. Variants of CCA include regularized CCA \cite{vinod1976reg_cca}, Kernel-CCA (KCCA) \cite{akaho2006kcca,yu2014kcca,bach2002kcca}, Nonparametric CCA (NCCA) \cite{michaeli2016ncca}, Deep-CCA (DCCA) \cite{andrew2013dcca}, randomized non-linear component analysis (RCCA) \cite{mineiro2014rcca} - a low rank approximation of KCCA, and Deep-Generalized-CCA (DGCCA) \cite{benton2017dgcca}. Although CCA-based methods provide a combined correlated representations, they suffer from scalability issues \cite{chandar2016correlational}. Another problem associated with CCA-based techniques is the fact that these methods tend to have poor performance for reconstruction of views. 

Another broad category into which the CRL techniques can be divided is autoencoder (AE) based approaches. AE are deep neural networks that try to optimize two objective functions \cite{ng2011auto,li2015auto}. The first objective is to find a compressed hidden representation of data in a low-dimensional vector space. The other objective is to reconstruct the original data from the compressed low-dimensional subspace. Multi-modal autoencoders (MAE) are two channeled AE that specifically performs two types of reconstructions \cite{ngiam2011multimodal}. The first is the self-reconstruction of view from itself, and the other is the cross-reconstruction where one view is reconstructed given the other. These reconstruction objectives provide MAE the ability to adapt towards transfer learning tasks as well. One recently proposed variant of MAE is Correlation Neural Networks (CorrNet) \cite{chandar2016correlational} which presents an improvement to MAE by introducing a correlation term in the objective function that tries to maximize the correlation between the hidden representations of different views. Some limitations of the CorrNet includes usage of the simple neural layer for encoding and decoding and using the final hidden representations in the correlation loss function.

Deep networks have grabbed the attention of many ever since the advent of state-of-the-art results using CNN networks. Apart from that, a lot of techniques have been worked upon to improve the classical CNN models. Regularization methods to reduce over-fitting, activation functions and modified convolution layers are some of them. Dropout technique \cite{hinton2012improving_dropout} is a very commonly used stochastic regularization technique which is also being used in our model. For activation function, sigmoid functions are avoided as they have a problem of vanishing gradient when large networks are involved. Hence, we use the rectified linear unit (ReLU) \cite{nair2010rectified} which is mathematically efficient. Batch normalization \cite{ioffe2015batch} has been used to increase the training rate providing us with better correlation values as compared to the previous papers. It is a stabilizing mechanism for training a neural network by scaling the output of hidden layers to zero norm and unit variance. This process of scaling reduces the change of distribution between neurons throughout the network and helps to speed up the training process.

In this paper, we focus our attention on the improvement in objective function of CorrNet, thereby enhancing the learned joint representations and reconstruction of views as well. Specifically, main contributions of the paper are
\begin{enumerate}
\item We introduce convolution layers to the encoding phase of Correlation multi-modal CNN (CorrMCNN), and deconvolution is used in the decoding stage.
\item We use batch normalization in the intermediate layers of proposed model along with tied weights architecture.
%
%
\item Instead of using final hidden representations in the correlation loss, we enforce correlation computation at each intermediate layer. We further experiment with the reconstruction of hidden representation at every individual step.
\end{enumerate}

The rest of the paper is organized as follows. In section \ref{sec_rel} we discuss previous work related to our work which is followed by the discussion of CorrMCNN in Section \ref{sec_tech}. The experimental details and results are shown in section \ref{sec_exp} and then our work is concluded in section \ref{sec_con}.

\section{Related Work}
\label{sec_rel}
Several methods have been used for generating the hidden representation of two views of data. This hidden representation can also be used to construct the other missing view when one of the views is provided as an input. CCA \cite{hardle2007canonical,thompson2005canonical} is the oldest model which is able to achieve this feat but has the drawbacks of scalability to large datasets. Scalability has been achieved by \cite{lu2014scalable_cca} but at the cost of performance. Also, CCA results in low-quality reconstruction when one view is reconstructed from the other. Finally, CCA suffers from a severe disadvantage of not being able to be used in several real world problems, as it cannot benefit from additional non-parallel, single-view data. Several variants of CCA has been developed ever since it's importance grew in the field of computer vision. These modifications include Regularized CCA \cite{mineiro2014rcca}, KCCA \cite{akaho2006kcca,yu2014kcca} and a recently emerged low-rank approximation of KCCA known as RCCA \cite{mineiro2014rcca}. However, CCA is restricted to linear projections, while KCCA works only on a fixed kernel. 

Several deep learning models have been proposed that aim at solving the above problems. Deep canonical correlation analysis (DCCA) \cite{andrew2013dcca} works for pairs of inputs through two network pipelines and evaluates the results of each pipeline via the CCA loss.
CorrNet \cite{chandar2016correlational} and deep canonically correlated autoencoders (DCCAE) \cite{wang2015deepcca} are an extended version of the concept of auto-encoder that take two input views and produce two output views. DCCAE is an extension of DCCA which takes into account self-reconstruction and correlation but doesn't consider cross-reconstruction. MAE \cite{ngiam2011multimodal} is another method used for CRL with architecture similar to CorrNet. The only difference exists between MAE and CorrNet is in their objective functions. Unlike MAE, CorrNet's objective function enforces the model to learn the correlated common representations as well. Also, CorrNet aims to minimize all the error at once, unlike its predecessor which adopts a stochastic version. 


\section{Technique Used}
\label{sec_tech}

\begin{figure*}
\centering\includegraphics[width=1\textwidth,height=0.26\textheight]{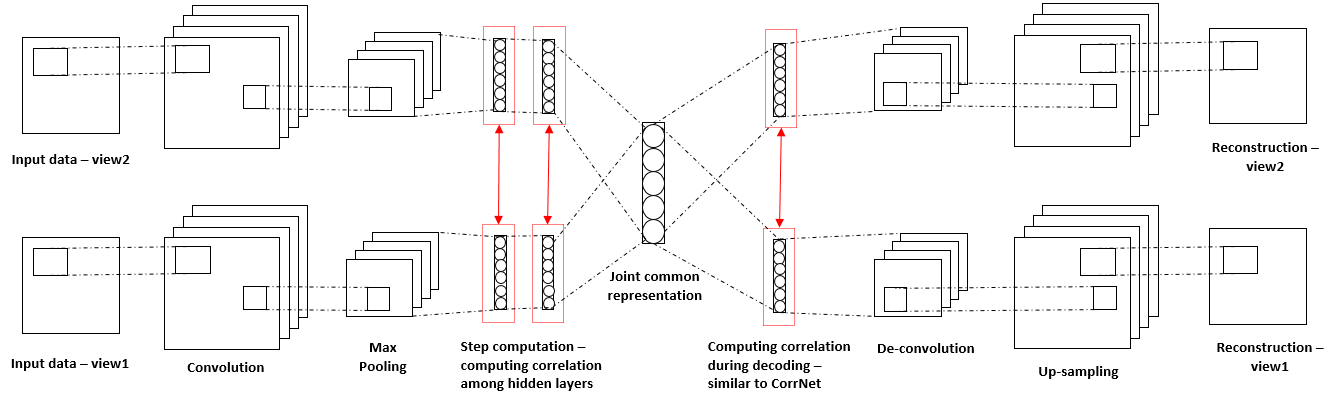}
\caption{Overview of the CorrMCNN. The bidirectional arrows shows the step correlation computation and cross-reconstructions at the intermediate steps.}
\label{fig:one}
\end{figure*}

\subsection{Convolution Auto-encoder}
Convolution auto-encoder (CAE) \cite{masci2011cae} is novel idea for unsupervised feature learning that are a variant of convolution neural network (CNN). However, the difference lies in the fact that CNNs are usually referred to as supervised learning algorithms. Usually, an unsupervised pre-training is done greedily and with the help of layers, and after that, the weights are fine-tuned using back-propagation. One limitations of fully connected AEs and DAEs when used for multi-modal learning is that they both ignore the multi-dimensional image structure. This problem is not only concerned with ordinary sized inputs but further introduces extraneous parameters that force each feature to span the entire visual field. The main difference between CAEs and conventional AEs is that in CAEs the weights are shared everywhere in the input which preserves spatial locality. The variant of CAE that we use in this paper is shown in Figure \ref{fig:one}.

In Figure \ref{fig:one} the input data is passed on to two channels which add convolution and max-pooling to obtain a lower dimensional representation of the data. We add dense neural layers after the max-pooling so that the interaction between the hidden representations can be maximized. We further add dropout as regularization to these intermediate dense neural layers. Finally, the output of the dense neural layers is used to obtain a joint common representation. To reconstruct the original input from the joint common representations, the projections are passed through deconvolution and up-sampling. Here, we also add a dense neural layer between the joint common representations and the deconvolution. 

\subsection{CorrMCNN}
For learning a common representation of the two views of data, one of the views is to be reconstructed from the hidden representation. This task of reconstruction can be achieved by using a conventional auto-encoder. However, it is also required that one of the views can be predicted from the other view provided that both the representations are correlated. To achieve this functionality, multiple channeled autoencoders such as MAE and CorrNet are being used. The concept of correlation networks can be extended to allow for multiple hidden layers \cite{chandar2016correlational}. Multiple hidden layers ensure better interaction between the two views of the data as a result of passing the data through more non-linearity.

For training the proposed model, we target the following goals in our objective function:
\begin{itemize}
  \item Minimize the self-reconstruction error.
  \item Minimize the cross reconstruction error at each intermediate step.
  \item Introduce batch-normalization at the intermediate dense neural layers.
  \item Maximize the correlation between the hidden representation of both views at each encoding step.
\end{itemize}

Given the input as $z_i =\{x_i; y_i\}$, where $z_i$ is the concatenated representation of input views $x_i$ and $y_i$ (corresponding to $view1$ and $view2$ in Figure \ref{fig:one}), the self and cross-reconstruction losses are defined as
\begin{flalign}
L_1 &= \sum_{i=0}^{N} L(z_i,g(h(z_i)) \\
L_2 &= \sum_{i=0}^{N} L(z_i,g(h(x_i)) \\
L_3 &= \sum_{i=0}^{N} L(z_i,g(h(y_i)) \\
L_4 &= \sum_{k=0}^{K}\sum_{i=0}^{N} L(h(x_i^k),h(y_i^k)) \\
L_5 &= \sum_{i=0}^{N} L(g(h(x_i)),g(h(y_i))
\end{flalign}
where, $g$, $h$ are non-linearities generally taken as sigmoid or relu, $g(h(x_i^k))$ and $g(h(y_i^k))$ are the hidden representations at the $k^{th}$ intermediate hidden layer (In Figure \ref{fig:one} the value of $k$ is 2), and $L$ is the mean square error function. Step computation (shown in Figure \ref{fig:one}) is given by Equation 4. The reason for introducing step computation is the fact that the hidden representations share similarities and thus reconstruction of one hidden representation from the other helps the model in final reconstruction of views, improving interaction between layers as well. In losses $L_2$ and $L_3$ (used for cross reconstruction), $x_i$ and $y_i$ are computed using a $0$-vector in place of the other view.

Finally, to maximize the interaction between the two views correlation loss is added as 
\begin{align}
L_6 &= \lambda\ corr(h(X),h(Y))\\
L_7 &= \sum_{k=0}^{K} {\lambda}_{k}\ corr(h(X^k),h(Y^k))
\end{align}
where $h(X)$ and $h(Y)$ are the projections from the combined model (projection from joint common representation in Figure \ref{fig:one}), with $X$ and $Y$ being the representations of input view obtained after passing through convolution and pooling layers, and ${\lambda}_{k}$ are the correlation regularization hyper-parameter used for each intermediate ${k}^{th}$ encoding step (similarly $\lambda$ is used in the decoding phase). 

Equation 7 is the step-wise computation of correlation between the corresponding hidden views. 
CorrNet uses $L_6$ to increase the correlation among two views while CorrMCNN uses combined correlation losses computed at each intermediate hidden layer (In Figure \ref{fig:one} the value of $k$ is 2 during encoding and 1 in decoding).

Finally, the CorrMCNN is optimized using the following objective function using $adam$ as optimizer 
\begin{dmath}
L(\theta) = \sum_{i=0}^{5} {L}_{i} - \sum_{j=6}^{7} {L}_{j}
\end{dmath}
where $\theta$ are the parameters of CorrMCNN. Here, we minimize the self-reconstruction and cross-reconstruction whereas the correlation between the views is maximized.

\subsection{Batch Normalization}
To maximize the correlation, the variance of every neuron's output must be increased. This can be implemented by introducing batch normalization (BN) \cite{ioffe2015batch} layers. These BN layers alleviate the loss due to variance by enforcing unit variance and eradicating the impact of weights of the hidden layers on the output's variance. This method allows us to achieve the same accuracy in fewer training steps. Apart from that, BN layers also allow high learning rates. Traditionally, high learning rates in deep neural networks resulted in vanishing gradient and exploding gradient problems, as well as the issue of getting struck at poor local minima. Batch Normalization proves out to be quite beneficial in such cases. We use BN layers in the dense neural layers during encoding and decoding.

\section{Experiments}
\label{sec_exp}

\subsection{Datasets}
All the models mentioned in Section \ref{sec_rel} are trained on MNIST dataset, and X-Ray Microbeam Speech data (XRMB) \cite{westbury1990xrmb}, and the sum correlation along with the transfer learning accuracy is calculated. MNIST handwritten digits dataset has 60,000 images of handwritten digits for training and 10,000 for testing. Each image is split vertically into two halves so as to obtain an image of 28 x 14 or in other words, 392 features for each view of data. A 50-dimensional joint common representation space has been employed in this case. 

XRMB dataset contains simultaneous acoustic and articulatory recordings. The acoustic features are MFCCs \cite{logan2000mfcc} for the given frames, resulting in a 273-dimensional vector per unit time. On the other hand, articulatory data is represented as an 112-dimensional vector. For benchmarking, 40,000 samples are used for training, and 10,000 samples are used for testing purposes. XRMB dataset has a joint common representation of 112 dimensions.

%
\subsection{Compared Methods}
We compare CorrMCNN with both CCA based and AE based approaches. For canonical based approaches we perform comparisons with CCA \cite{hardle2007canonical}, DCCA \cite{andrew2013dcca}, KCCA \cite{akaho2006kcca}, RCCA \cite{mineiro2014rcca} and DCCAE \cite{masci2011cae}. Apart from CCA based approaches we have also compared CorrMCNN with recently proposed 2-way nets \cite{eisenschtat2016linking_2way}.\footnote{Results of KCCA, MAE and CorrNet-Org on MNIST dataset are taken from \cite{chandar2016correlational}. Code for CorrNet-Org by \cite{chandar2016correlational} is available at :- https://github.com/apsarath/CorrNet }

For AE based approaches we perform comparisons with  MAE \cite{ngiam2011multimodal}, CorrNet-Org \cite{chandar2016correlational} and CorrNet-Mod \cite{eisenschtat2016linking_2way} \footnote{CorrNet-Mod - Revised results of CorrNet on MNIST dataset are stated by \cite{eisenschtat2016linking_2way}}.

\subsection{Experimental setup}
We have used 2 different architectures of CorrMCNN - one with $L_7$ loss term and other without $L_7$ term:
\begin{enumerate}
\item CorrMCNN-arc1 - using convolution layers and 2 dense neural layers with $k$=2 during encoding. For decoding, one dense layer ($k$=1) is used for reconstruction with dropout of 50 \%. $L_7$ loss term is excluded in this architecture with $\lambda$ = 0.02 in $L_6$. 
\item CorrMCNN-arc2 - using convolution layers and 2 dense neural layers with $k$=2 during encoding. For decoding we use deconvolution layers along with one dense layer ($k$=1) for reconstruction and $L_7$ loss term is included in this architecture. Values of ${\lambda}_{1}$=0.005 and ${\lambda}_{2}$=0.01 during encoding and for decoding $\lambda$ = 0.005 in $L_7$ term. For $L_6$, $\lambda$ = 0.02 is used.
\end{enumerate}
 Linear SVM implementation as mentioned in \cite{pedregosa2011scikit} is used for transfer learning along with a 5-fold cross validation on images from the MNIST half matching dataset.
 
\subsection{Results}
\label{sec_res}

\begin{table}[h!]
\centering
\begin{tabular}{|l|l|l|}
 \hline
 \textbf{Model} & {\textbf{MNIST}} & {\textbf{XRMB}} \\
 \hline
 KCCA \cite{yu2014kcca} & $30.58$ & $N.A.$\\
 DCCA \cite{andrew2013dcca} & $39.7$ & $92.9$\\
 MAE \cite{ngiam2011multimodal} & $24.40$ & $N.A.$\\
 RCCA \cite{mineiro2014rcca} & $44.5$ & $104.5$\\
 DCCAE \cite{masci2011cae} & $25.34$ & $41.47$\\
 Reg-CCA \cite{vinod1976reg_cca} & $28.0 $ & $16.90$\\
 CorrNet-Org \cite{chandar2016correlational} & $47.21$ & $81.54$\\
 CorrNet-Mod & $48.07$ & $95.01$\\
 2WayNet \cite{eisenschtat2016linking_2way} & $49.15$ & $\textbf{110.18}$ \\
 \hline
 CorrMCNN-arc1& $49.08$ & $105.04$\\
 CorrMCNN-arc2& $\textbf{49.33}$ & $105.59$\\
 \hline
\end{tabular}
\bigskip
\caption{Sum correlation captured in the joint common representations
learned by different models on MNIST and XRMB dataset.}
\label{tab_res1}
\end{table}

\begin{table}
\centering
\begin{tabular}{|l|l|l|}
 \hline
 \textbf{Model} & {\textbf{Left to Right}} & {\textbf{Right to Left}} \\
 \hline
 CCA \cite{vinod1976reg_cca} & $65.73$ & $65.44$\\
 KCCA \cite{yu2014kcca} & $68.1$ & $75.71$\\
 DCCA \cite{andrew2013dcca} & $70.06$ & $72.43$\\
 MAE \cite{ngiam2011multimodal} & $64.14$ & $68.88$\\
 CorrNet-Org \cite{chandar2016correlational} & $77.05$ & $78.81$\\
 CorrNet-500-50 \cite{chandar2016correlational}& $80.46$ & $80.47$ \\
 \hline
 CorrMCNN-arc1 & $\textbf{90.27}$ & $\textbf{91.16}$\\
 CorrMCNN-arc2 & $84.23$ & $85.76$\\
 \hline
 Single view & $81.62$ & $80.06$ \\
 \hline
\end{tabular}
\bigskip
\caption{Transfer learning accuracy using the learned joint common representations on MNIST dataset.}
\label{tab_res2}
\end{table}

For evaluating the performance of the proposed model, we have used two evaluation metrics: total sum correlation and transfer learning (reconstruction of one view of MNIST dataset using the other). These results are shown in Table \ref{tab_res1} and \ref{tab_res2} respectively.

As shown in Table \ref{tab_res1}, CorrMCNN achieves the highest sum correlation value for MNIST dataset while achieving the second best on XRMB dataset. The introduction of $L_7$ loss term that is the step based correlation achieves the highest correlation for the MNIST dataset. One possible reason for slightly lower performance of CorrMCNN on XRMB dataset is the fact that it consists of 1D MFCC features that are the compressed representation of original acoustic signals. Despite the 1D nature of XRMB, CorrMCNN achieves comparable performance on the task of learning correlated joint common representations.

\begin{figure}
\centering\includegraphics[width=0.45\textwidth,height=0.16\textheight]{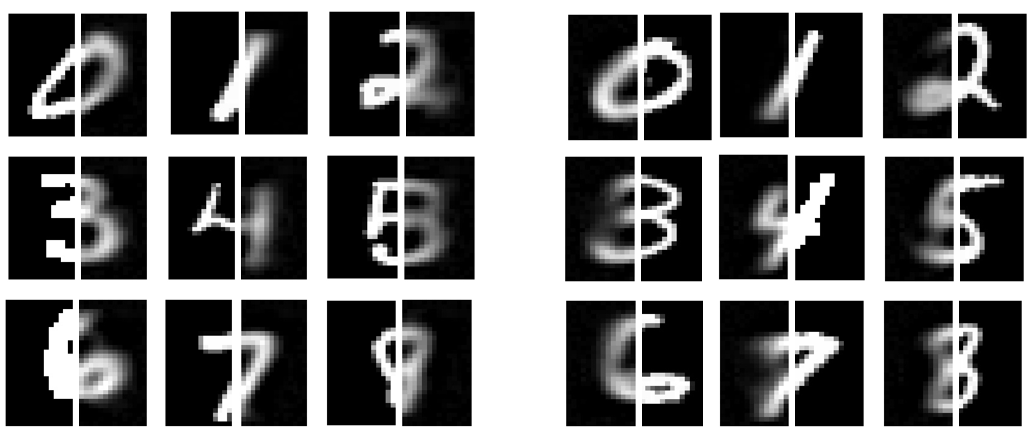}
\caption{Reconstruction of right view from left and left view from right by CorrMCNN on MNIST dataset.}
\label{fig:two}
\end{figure}

The results of transfer learning task is shown is shown in Table \ref{tab_res2}. Here, $Single\ view$ corresponds to the SVM trained and tested on the single view. Both architectures of CorrMCNN are able to achieve better performance than the current state-of-the-art techniques. The increment in the cross-reconstruction accuracy is more than $10\%$ than the $Single\ view$ which acts as the baseline (ideal case) for this task. Here, CorrMCNN-arc1 performs better than its arc2 counterpart and the reason is the introduction of $L_4$ and $L_5$ cross-reconstruction loss terms. We observe that the introduction of $L_7$ term in the loss function slightly over-fits on the task of cross-reconstruction which is the reason for lower accuracy. Finally, as an illustration the results for reconstruction of either of the views is shown in Figure \ref{fig:two}. 

During experimentations we observed that the introduction of batch normalization helps the proposed model to converge faster giving high correlation values. The only downside of BN is the slight drop in reconstruction accuracy during transfer learning. Deconvolution and up-sampling helps CorrMCNN to increase the sum correlation. Finally, the introduction of $L_4$ and $L_5$ terms helps in improving the reconstruction accuracy, bolstering the overall cross-reconstruction.

\section{Conclusion}
\label{sec_con}

In this paper, we proposed CorrMCNN which learns common representations for multi-view data. The introduction of step correlation terms as introduced in Section 3 (loss terms $L_4$, $L_5$ and $L_7$) helps to achieve a highly correlated common representation. The proposed model not only increases the interaction between the representations but helps in improvement of self and cross-reconstruction (shown in Table \ref{tab_res2}). The results shown in Table \ref{tab_res1} and Table \ref{tab_res2} shows that the proposed model is able to 
achieve state-of-the-art performance on either of the benchmark datasets on common representation learning and transfer learning tasks. 

On final remarks, we observed some limitations of AE and canonical correlation-based approaches when used on high dimensional data such as 3D images and video frames. These models show state-of-the-art performance on increasing the correlation between the two views but the cross reconstruction of views is not possible. We speculate that improving the cross reconstruction accuracy will have a positive effect on computing a joint common representation task as either of these are dependent on each other. We intend to work in this direction in the near future. 

\section*{Acknowledgement}
We would like to express our gratitude towards Institute Compute Center (ICC), Indian Institute of Technology Roorkee, for providing us with necessary resources for this paper.


\bibliographystyle{IEEEtran}
\bibliography{egbib}
%



%
\end{document}